\documentclass[journal]{IEEEtranTIE}
\usepackage{amsmath,amsfonts}
\usepackage{algorithm}
\usepackage{array}
\usepackage[caption=false,font=footnotesize]{subfig}
\usepackage{textcomp}
\usepackage{stfloats}
\usepackage{url}
\usepackage{verbatim}
\usepackage{float}
\usepackage{xcolor}
\usepackage{graphicx}
\usepackage{algpseudocode}
\usepackage{amsmath}
\usepackage{cite}
\usepackage{multirow}
\usepackage{multicol}
\usepackage{graphicx}
\usepackage{booktabs}
\usepackage{cite}
\usepackage{siunitx}
\usepackage[bookmarks=true]{hyperref}
\hypersetup{hidelinks,
            colorlinks=true,
            }
\usepackage{silence}
\usepackage[commandnameprefix=ifneeded]{changes}
\usepackage{todonotes}

\usepackage{hyperref}

\begin{document}

\title{DRL-TH: Jointly Utilizing Temporal Graph Attention and Hierarchical Fusion for UGV Navigation in Crowded Environments}

\author{Ruitong Li, Lin Zhang, Yuenan Zhao, Chengxin Liu, Ran Song, and Wei Zhang
        
}

\maketitle

\begin{abstract}
Deep reinforcement learning (DRL) methods have demonstrated potential for autonomous navigation and obstacle avoidance of unmanned ground vehicles (UGVs) in crowded environments. Most existing approaches rely on single-frame observation and employ simple concatenation for multi-modal fusion, which limits their ability to capture temporal context and hinders dynamic adaptability.
To address these challenges, we propose a DRL-based navigation framework, DRL-TH, which leverages temporal graph attention and hierarchical graph pooling to integrate historical observations and adaptively fuse multi-modal information. Specifically, we introduce a temporal-guided graph attention network (TG-GAT) that incorporates temporal weights into attention scores to capture correlations between consecutive frames, thereby enabling the implicit estimation of scene evolution. In addition, we design a graph hierarchical abstraction module (GHAM) that applies hierarchical pooling and learnable weighted fusion to dynamically integrate RGB and LiDAR features, achieving balanced representation across multiple scales. 
Extensive experiments demonstrate that our DRL-TH outperforms existing methods in various crowded environments. We also implemented DRL-TH control policy on a real UGV and showed that it performed well in real-world scenarios. 
\end{abstract}


\section{Introduction}
\IEEEPARstart{F}{or} crowded scenarios, effective navigation requires unmanned ground vehicles (UGVs) to exhibit efficient multi-modal perception and obstacle avoidance to ensure both safety and task completion. While classical navigation methods have been widely applied to UGV navigation \cite{wang2023switching-tie, cao2024, shen2019observability-tie}, they requires maintaining an extensive “if-else” case repository, leading to limited scalability and insufficient fault tolerance. Consequently, recent studies have increasingly shifted toward learning-based approaches that leverage data-driven models to enhance UGV navigation in such settings.

There exist some works based on deep learning (DL) \cite{zhao2025rakd, zhai2020robust-tie, lrt2025il} and imitation learning (IL) \cite{codevilla2018end, pfeiffer2017perception, abdou2019end} to address the limitations of classical methods in crowded environments. DL methods quintessentially focus on environmental perception tasks, such as object detection and semantic segmentation, and do not explicitly learn the navigation policy. Some studies \cite{qiu2017using, patel2019deep, lrt2025il} have attempted to train policy using offline annotations from real-world environments. Such policy annotations are not only time-consuming and laborious to generate, particularly at a large scale in highly dynamic environments, but also subject to a fixed, limited, and discrete set of action states. Similarly, IL methods depend on expert datasets to clone expert behavior, enabling UGVs to emulate demonstrated trajectories. This approach often fails to adequately evaluate the quality of navigation behavior, and the trained policy can not generalize beyond the scope of the expert data \cite{codevilla2019exploring}. When the expert dataset is insufficient in size and diversity, IL methods are prone to overfitting and encounter difficulties in transferring from simulation to real-world scenarios (\textit{Sim2Real}).

In contrast, deep reinforcement learning (DRL) learns optimal navigation policies by maximizing expected returns through extensive interactions with the environment. Prior work \cite{huang2021ral, liang2018cirl, zeng2024poliformer, chen2020relational} maps raw sensor inputs to navigation commands, enabling UGVs to adapt to dynamic conditions. However, extracting useful latent features from high-dimensional raw data places a significant learning burden on policy models, often requiring larger and more complex networks. Other studies \cite{zhang2021coach, xie2023drl-vo, xue2023combining, xieyd2025iros} preprocess sensor data using perception networks to generate intermediate-level feature maps, thereby reducing complexity and enhancing the feasibility for \textit{Sim2Real} transfer. Although these maps are easier to process, modality fusion is commonly performed through concatenation and lacks dynamic adaptability. Recently, Transformer-based methods \cite{zeng2024poliformer, huang2023goaltransformer, xu2024transformer} offer higher representation capacity, and they are sensitive to geometric or semantic misalignments between modalities, which may lead to feature shifts and make UGV decisions unstable. In addition, effective obstacle avoidance requires the navigation policy to understand the actual context of the scene in crowded environments. Most current works focus on instance-level prediction \cite{zhao2025rakd, zeng2025navidiffusor, niedoba2023diffusion}, such as trajectory forecasting or diffusion model inference. While these methods can achieve high accuracy, they are computationally intensive and thus unsuitable for deployment on UGV platforms with limited computing resources.

To address the aforementioned challenges, this paper incorporates graph neural networks (GNNs) for dynamic integration of multi-modal features and implicit context reasoning, serving as an alternative to instance-level prediction and enabling deployment in real-world environments. In particular, we propose DRL-TH, a deep reinforcement learning framework that integrates GNNs for UGV navigation in crowded environments. DRL-TH initially processes raw LiDAR and RGB data through feature extraction networks, representing them as delineations of traversable and non-traversable areas in bird's-eye view (BEV) and first-person view (FPV). These features are then sequentially weighted and fused by the temporal-guided graph attention network (TG-GAT) and graph hierarchical abstraction module (GHAM), ultimately feeding into a proximal policy optimization (PPO) \cite{schulman2017ppo} network to output navigation commands.

Specifically, we first represent a frame sequence as a graph, where nodes correspond to frames and edges capture temporal relationships. TG-GAT incorporates temporal weights into attention scores to model these relationships, enabling the UGV to effectively capture of scene evolution. GHAM then utilizes hierarchical pooling and learnable weighted fusion to dynamically integrate RGB and LiDAR features. GHAM employs learnable weights to balance modality features across multiple scales, ensuring adaptive fusion optimized for environmental complexity. We conducted extensive experiments in both simulated and real-world environments. The results demonstrate that DRL-TH achieves superior navigation performance in crowded scenarios with various obstacle densities, effectively avoiding both static and dynamic obstacles.

In summary, the contributions of this work are threefold:
\begin{enumerate}
    \item We design DRL-TH, a DRL framework for UGV navigation in crowded environments that integrates historical observations and adaptively fuses multi-modal information. The framework incorporates two key modules: TG-GAT and GHAM.
    \item TG-GAT incorporates temporal weights into attention scores to effectively capture temporal correlations between consecutive frames. GHAM employs hierarchical pooling and learnable weights to dynamically integrate and balance the features across multiple scales.  
   \item We evaluate DRL-TH on a UGV in various simulated and real-world environments, demonstrating its superiority in terms of safety and efficiency.
\end{enumerate}

\section{Related Work}
This section reviews two research areas related to our study: reinforcement learning and graph neural networks for navigation.

\subsection{Reinforcement Learning for Navigation}  
RL trains agents to maximize cumulative rewards by exploring actions and receiving feedback from environmental dynamics. This trial-and-error paradigm is suited for dynamic and unstructured settings, as it allows robots to adapt to unforeseen obstacles and multi-agent interactions without predefined rules. Many RL methods output driving actions based on raw sensor inputs. Huang et al. \cite{huang2021ral} utilized PPO \cite{schulman2017ppo} to construct a dual-branch policy network that processes image and laser inputs, outputting linear and angular velocities to enable UGV navigation and obstacle avoidance. Liang et al. \cite{liang2018cirl} presented an off-policy replay-memory-based RL algorithm, processing raw image inputs to generate specific control actions for autonomous navigation tasks. Zeng et al. \cite{zeng2024poliformer} proposed a transformer-based framework that employs on-policy RL to process raw RGB inputs for finding preset objects in indoor scenes, utilizing two different transformers to support extended memory and reasoning. While RL methods processing raw sensor inputs enable direct action mapping for navigation, the high dimensionality of these inputs entails larger networks and increases computational demands, prompting many methods to leverage intermediate-level feature representations for more efficient policy learning. For example, Zhang et al. \cite{zhang2021coach} designed the privileged information into BEV images to train an RL framework for autonomous navigation tasks. Xie et al. \cite{xie2023drl-vo} developed a pedestrian kinematic map representation that encodes the relative positions and velocities of detected pedestrians within a grid-based structure, enabling robot navigation in dynamic pedestrian environments. Xue et al. \cite{xue2023combining} introduced a DRL policy, which progressively generates a series of local landmarks using the current depth map observations. These approaches utilize the low-dimensional representations to simplify the observation space, enhancing RL efficiency in dynamic environments.
\subsection{Graph Neural Networks for Navigation}
Graph Neural Networks (GNNs) are a class of neural networks designed to learn functions on graph-structured data \cite{velivckovic2017gat, ying2018hierarchical}, with nodes representing objects and edges denoting their relations. Object dynamics can be encoded in node update functions, while interaction dynamics can be encoded in edge update functions. GNNs are primarily applied in two areas for navigation. First, environment modeling formulates navigation environments as graphs, with nodes as obstacles or robots and edges capturing spatial or semantic relations \cite{chen2020relational, ye2022multi}. Chen et al. \cite{chen2020relational} proposed a graph learning approach to predict human motion, enabling multi-step lookahead planning for robots. Ye et al. \cite{ye2022multi} integrated GNNs with DRL to navigate a swarm of unmanned aerial vehicles (UAVs) based on local observations. Second, the path planning model path points as nodes and reachability as edges, predicting traversability probabilities. Chen et al. \cite{chen2020autonomous} combined GNNs with DRL to enable robots to learn exploration strategies in unknown environments without human intervention. \cite{chen2019behavioral} and \cite{kahn2018self} developed a graph localization network for visual navigation tasks. In contrast to instance-based environment modeling, we leverage GNNs within DRL-based navigation to enable the implicit learning of scene evolution, where nodes represent features from different frames and edges represent frame-wise feature similarities and temporal proximities.

\section{Method}
\begin{figure*}[t]
\centering
\includegraphics[width=\textwidth]{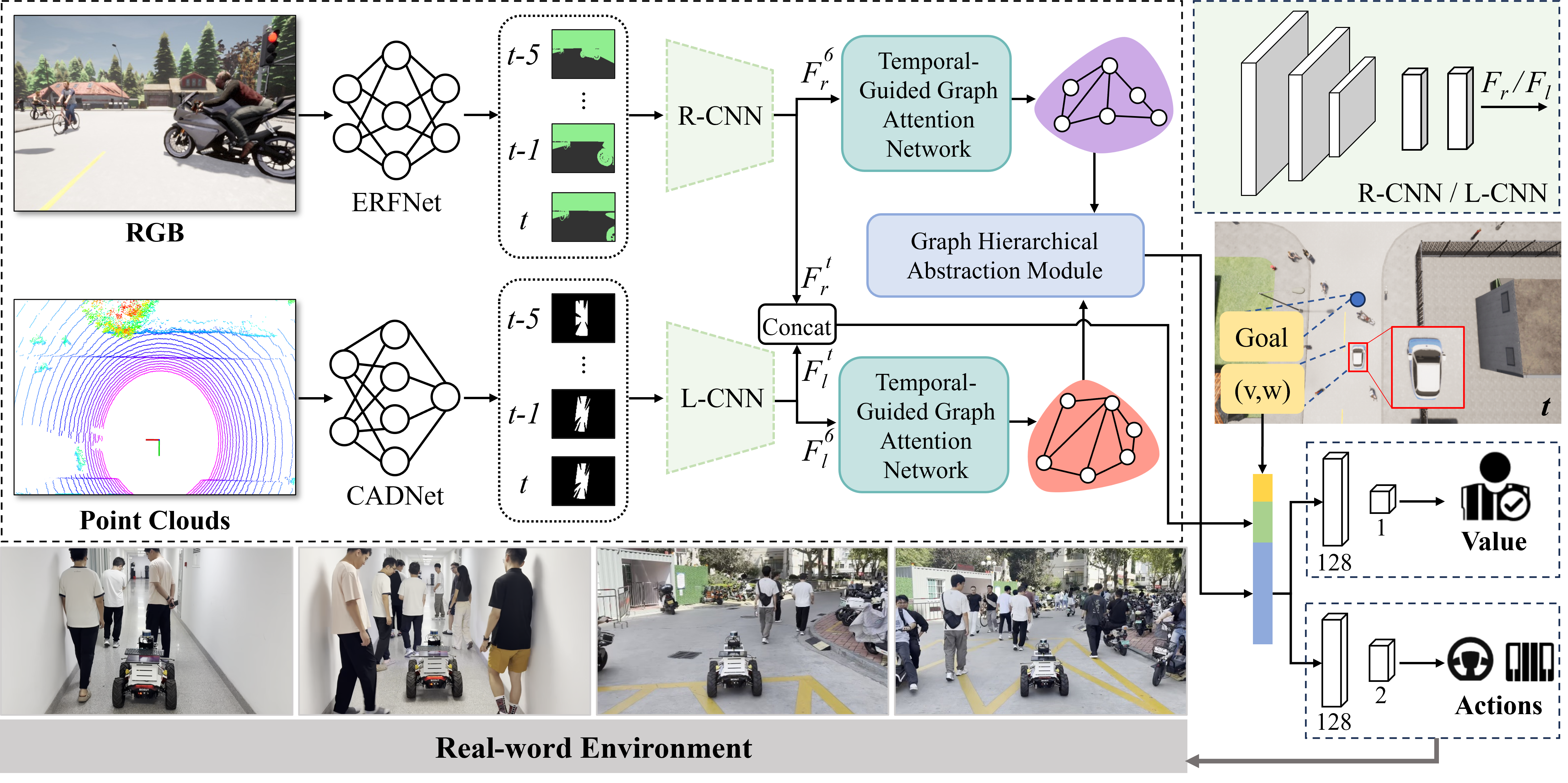}
\caption{Overall of our DRL-TH framework, where the detailed architectures of the temporal-guided graph attention network (TG-GAT) and graph hierarchical abstraction module (GHAM) are illustrated in Figs. \ref{TGGAT} and \ref{diffpool}.}
\label{pipline}  
\end{figure*}

In this section, we propose a multi-modal fusion framework based on DRL for UGV navigation. The overall architecture is illustrated in Fig. \ref{pipline}. First, the framework converts raw RGB images and LiDAR point clouds into segmentation maps of traversable and non-traversable areas under FPV and BEV perspectives by ERFNet \cite{romera2017erfnet} and CADNet \cite{xie2023cad} from the current and previous five frames, respectively. 
These maps are encoded by two CNNs with different architectures to extract features, which are then processed by TG-GAT to capture the temporal continuity across consecutive frames and by GHAM to achieve dynamic fusion of the two types of features.
The fused features are then concatenated with the current frame RGB and LiDAR features, reduced in dimensionality via an MLP, and combined with UGV velocity and target information. These features are processed by the RL-based policy to generate control commands. The final outputs are the linear velocity and the angular velocity which will be fed into the UGV controller.

\subsection{Reinforcement Learning for UGV Navigation}
\subsubsection{Algorithm Selection}

Since UGV navigation requires efficient decision-making in continuous action spaces, we adopt PPO for training the agent due to its stable and efficient learning performance. PPO optimizes a clipped surrogate objective function using gradient ascent, which is formulated as follows
\begin{equation}
\mathcal{L}^{CLIP}(\theta) = \mathbb{E}_t \left[ \min \left( r_t(\theta) A_t, \text{clip} (r_t(\theta), 1 - \epsilon, 1 + \epsilon) A_t \right) \right]
\end{equation}
where \( r_t(\theta) = \frac{\pi_\theta (a_t | s_t)}{\pi_{\theta_{old}} (a_t | s_t)} \) represents the probability ratio between the current policy \(\pi_\theta\) and the old policy \(\pi_{\theta_{old}}\), $A_t$ represents the advantage function at timestep $t$, and $\epsilon$ is set to 0.1. The \(\text{clip}(\cdot)\) function constrains the value of \(r_t(\theta)\) within the range \([1 - \epsilon, 1 + \epsilon]\), limiting the extent of policy updates and stabilizing training.

\subsubsection{Observation Space and Action Space}
In our task, the raw RGB images and LiDAR point clouds are high-dimensional, which can lead to excessive computational cost, slow convergence, and poor policy generalization. To address these issues, we pre-process the raw data to reduce computational complexity, filter out redundant information, and improve the generalization ability of the RL model, as detailed in Section~\ref{network part}. The observation $o_t$ consists of four components: the image $o_t^c$ from the camera, which providing the traversable area from the FPV, the image $o_t^l$ derived from LiDAR, which representing the traversable area from the BEV, the UGV’s current velocity $v_t$ and angular velocity $w_t$, and the position of the goal $g_t$. In practical applications, when the goal is distant, we introduce intermediate sub-goals during the path-planning process. This strategy helps to accelerate training and improve navigation efficiency by guiding the UGV through structured waypoints.

The action space $a_t = [v_t, w_t]$ consists of two components: the forward linear velocity $v_t$ and the angular velocity $w_t$. The linear velocity is constrained within the range $[0,2]$ m/s, while the angular velocity is constrained within the range $[-1,1]$ rad/s.

\subsubsection{Reward Function}
\label{reward}
The RL-based strategy allows the UGV to make real-time adaptive decisions, enabling stable movement and successful task completion in complex environments. UGV needs to avoid collisions with static and dynamic objects, reduce unnecessary sudden stops, abrupt accelerations, or sharp turns, and reach the target within a limited time. Based on the above, we define the reward function as follows

\begin{equation}
\mathcal{R} = \mathcal{R_{\mathrm{success}}} + \mathcal{R_{\mathrm{safety}}} + \mathcal{R_{\mathrm{smooth}}}
\end{equation}         
where $\mathcal{R_{\mathrm{success}}}$ consists of $\mathcal{R_\mathrm{{d1}}}$, $\mathcal{R_\mathrm{{d2}}}$, and $\mathcal{R_{\mathrm{time}}}$. $\mathcal{R_\mathrm{{d1}}}$ is to encourage the agent to move forward steadily, expressed as

\begin{equation}
\mathcal{R_\mathrm{{d1}}} = 1 \times (d_o - d_n)
\end{equation} 
where $d_o$ represents the recorded distance of the vehicle to the target point from the last update (updated every 10 time steps), and $d_n$ represents the current distance of the vehicle to the target point. If the agent moves significantly within a short period, it may lead to a sudden large decrease in distance, resulting in an abnormally high reward. To prevent this, when $|\mathcal{R_\mathrm{{d1}}}| > 0.9$, the reward is set to 0.

The purpose of $\mathcal{R_\mathrm{{d2}}}$ is to ensure the agent approaches the target point. A continuous reward function is used to gradually incentivize approaching behaviors.

\begin{equation}
\mathcal{R_\mathrm{{d2}}} = \max(0, 6 - 12d_g)
\end{equation} 
where $d_g$ represents the distance between the agent and the target point. 

To prevent the agent from remaining stationary, we assign a small constant penalty to $\mathcal{R_{\mathrm{time}}}$ (set to $-0.1$ in the experiments), which represents the standstill time penalty applied at each decision step.

$\mathcal{R_{\mathrm{safety}}}$ includes $\mathcal{R_\mathrm{{c}}}$ and $\mathcal{R_\mathrm{{b}}}$. When the agent collides with other static or moving objects, it receives a penalty

\begin{equation}
\mathcal{R_\mathrm{{c}}} =
\begin{cases} 
-10, & \text{if the agent collides} \\
0, & \text{otherwise}.
\end{cases}
\end{equation} 

$\mathcal{R_\mathrm{{b}}}$ is designed to prevent the vehicle from driving in the wrong direction. Completely reversing will result in a continuous penalty

\begin{equation}
\mathcal{R_\mathrm{{b}}} =
\begin{cases} 
-0.1, & \text{if the agent drives against traffic} \\
0, & \text{otherwise}.
\end{cases}
\end{equation} 

$\mathcal{R_{\text{smooth}}}$ is composed of $\mathcal{R_{\text{v}}}$, $\mathcal{R_{\text{w}}}$, and $\mathcal{R_{\text{u}}}$. $\mathcal{R_{\text{v}}}$ is to encourage the agent to maintain a relatively high linear velocity $v$. If $v$ is too low, the agent may remain stationary or move too slowly. Therefore, the reward function keeps it within a limited range

\begin{equation}
\mathcal{R_{\text{v}}} = 0.1 \times \max(0, v - 0.5).
\end{equation} 

$\mathcal{R_{\text{w}}}$ is defined to prevent the agent from making excessive steer. The penalty increases quadratically with angular velocity $w$ to avoid over-penalizing small angular velocities while applying greater penalties for large angular velocities

\begin{equation}
\mathcal{R_{\text{w}}} = -\min(0.5, 0.5 \times w^2).
\end{equation} 

$\mathcal{R_{\text{u}}}$ is introduced to penalize excessive linear acceleration $u$, encouraging smooth acceleration and deceleration. The penalty is quadratic to focus on large accelerations while minimizing impact on small ones, with an upper limit to prevent over-suppression of other rewards

\begin{equation}
\mathcal{R_{\text{u}}} = -\min(0.3, 0.1 \times u^2).
\end{equation}

\subsection{Network Architecture}
\label{network part}
Fig. \ref{pipline} illustrates our proposed end-to-end trainable framework, which predicts the linear and angular velocities of a UGV at the next timestep based on environment-aware data. Due to discrepancies between sensor data in the simulated environment and the real world, we designed intermediate-level feature maps for both camera and LiDAR modalities to achieve the \textit{Sim2Real} transfer.

RGB images from the camera are processed by a lightweight segmentation network, ERFNet, which converts raw RGB images into FPV representations of traversable and non-traversable areas. However, constrained by the limited onboard computational resources of UGVs and the scarcity of labeled data in real-world settings, ERFNet is difficult to generalize to complex environments. To address the above issue, we employ a teacher-student model to embed SegFormer \cite{xie2021segformer} into the training process of ERFNet. In the implementation, the SegFormer is pre-trained on labeled data from the Cityscapes dataset. Then, the unlabeled RGB images collected from the real world are fed into SegFormer to generate segmentation maps as pseudo-labels for training the ERFNet. Knowledge distillation between the teacher and student models further enhances the performance of ERFNet. The optimized ERFNet is applied to onboard image segmentation in both simulated and real environments, with segmentation outputs processed by the R-CNN module to generate the feature vector $F_r$.

Raw LiDAR point cloud data are processed by the CADNet, which converts them into BEV representations of traversable and non-traversable areas. CADNet is pre-trained in the CARLA simulator using a large dataset of directly acquired labeled data supplemented by a small set of manually labeled real-world data. The segmentation outputs are processed by the L-CNN module to generate the feature vector $F_l$.

Subsequently, feature vectors $F_{r}^{6}$ and $F_{l}^{6}$ (representing the 6 frames RGB and LiDAR features from $t-5$ to $t$, respectively) are fed into the TG-GAT to capture temporal continuity across consecutive frames, producing generate time-aware features $F_{rt}^{6}$ and $F_{lt}^{6}$. These features are dynamically fused by the GHAM to generate the multi-modal feature $F_{rl}$ across multiple scales. Then, $F_{rl}$ is concatenated with the current frame features $F_{r}^{t}$ and  $F_{l}^{t}$, forming a high-dimensional vector, which is reduced to a feature vector $F_{rl}^{'}$ via an MLP. $F_{rl}^{'}$ is further concatenated with the navigation goal and the current UGV’s linear and angular velocities (each encoded as feature vectors via fully connected layers) and input into a fully connected layer, yielding a 128×1 two-dimensional feature vector $F_z$. Finally, the $F_z$ is input into the Actions and Value modules, each comprising a single fully connected layer. The Actions module outputs linear and angular velocities, with Sigmoid and Tanh activation functions applied to constrain the output range. The Value module outputs a state value without an activation function, reflecting the environmental state evaluation.

\subsection{Temporal-Guided Graph Attention Network for Implicit Context Reasoning}

\begin{figure}[t]
\centering
\includegraphics[width=0.45 \textwidth]{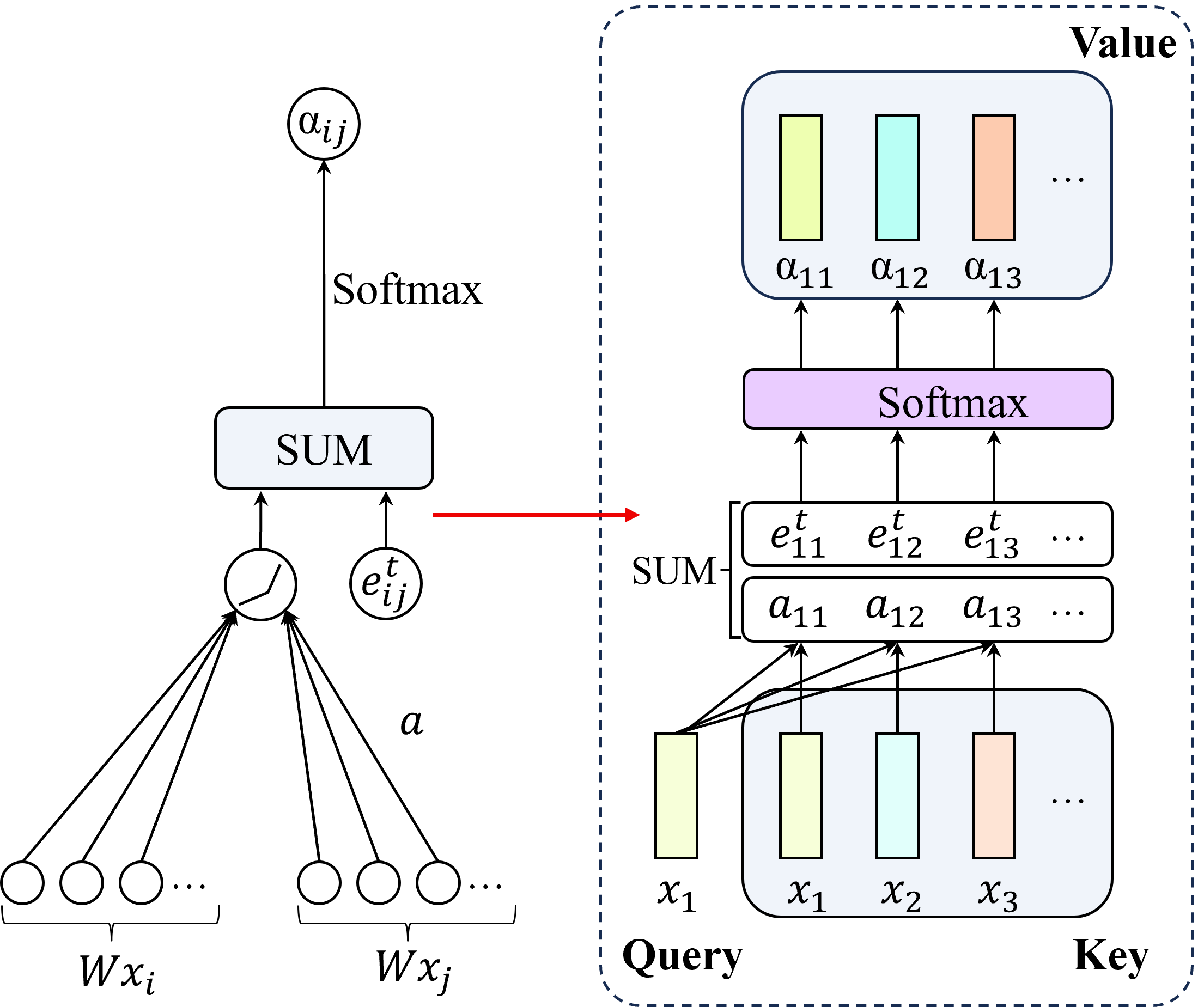}
\caption{Illustration of TG-GAT, incorporating temporal attention weights into the graph attention network framework.}
\label{TGGAT}  
\end{figure}

In this section, we demonstrate the temporal-guided graph attention network (TG-GAT), which is designed to capture temporal correlations in multi-frame observations for implicit learning of
scene evolution. TG-GAT models node relationships in graph-structured data by computing attention scores that integrate feature similarity and temporal weights, enabling adaptive aggregation of time-aware features from neighboring nodes, as illustrated in Fig. \ref{TGGAT}. 
Initially, each frame feature corresponds to a node in a sparse graph, with feature vectors $x_i, x_j$ derived from $F_r^6$ or $F_l^6$.

Edges are determined based on the cosine similarity between node features
\begin{equation}
\text{cos}(x_i, x_j) = \frac{x_i \cdot x_j}{\|x_i\| \|x_j\|}
\label{cossim} 
\end{equation}
where $\tau = 0.8$, an undirected edge $(i,j)$ is added if $\text{cos\_sim}(x_i, x_j) > \tau$. A shared linear transformation, parameterized by a \textit{weight matrix} $W\in \mathbb{R}^{128 \times 256}$, is applied to each node
\begin{equation}
z_i = W x_i.
\end{equation}

We compute \textit{attention coefficients} that combine feature similarity and temporal proximity
\begin{equation}
e_{ij} = \text{LeakyReLU}(a^T [z_i \| z_j]) + \beta e^{-\lambda |i-j|}
\end{equation}
where $a^T$ represents transposition of $a$, $[z_i \| z_j]$ denotes the concatenated vector, $a$ is a learnable attention vector, $\beta = 0.3$ scales the temporal weight, $\lambda = 0.5$ controls the decay of $\beta e^{-\lambda |i-j|}$ with frame interval $|i-j|$. This equation emphasizes frames closer to the current one because recent historical information is more relevant for scene evolution.

The attention coefficients are normalized via the Softmax function to obtain \textit{attention weights}
\begin{equation}
\alpha_{ij} = \frac{\exp(e_{ij})}{\sum_{k \in \mathcal{N}(i)} \exp(e_{ik})}
\end{equation}
where $\mathcal{N}(i)$ is some neighborhood of node $i$ in the graph, ensuring that the sum of $\alpha_{ij}$ equals 1. 

The node features are aggregated to produce updated representations
\begin{equation}
h_i = \text{ReLU}\left( \sum_{j \in \mathcal{N}(i)} \alpha_{ij} z_j \right)
\end{equation}
where $h_i \in \mathbb{R}^{256}$ is obtained by concatenating two attention heads, forming the output features $F_{rt}^6, F_{lt}^6 \in \mathbb{R}^{6 \times 256}$.

\subsection{Graph Hierarchical Abstraction Module for Multi-modal Fusion}

\begin{figure}[t]
\centering
\includegraphics[width=0.45 \textwidth]{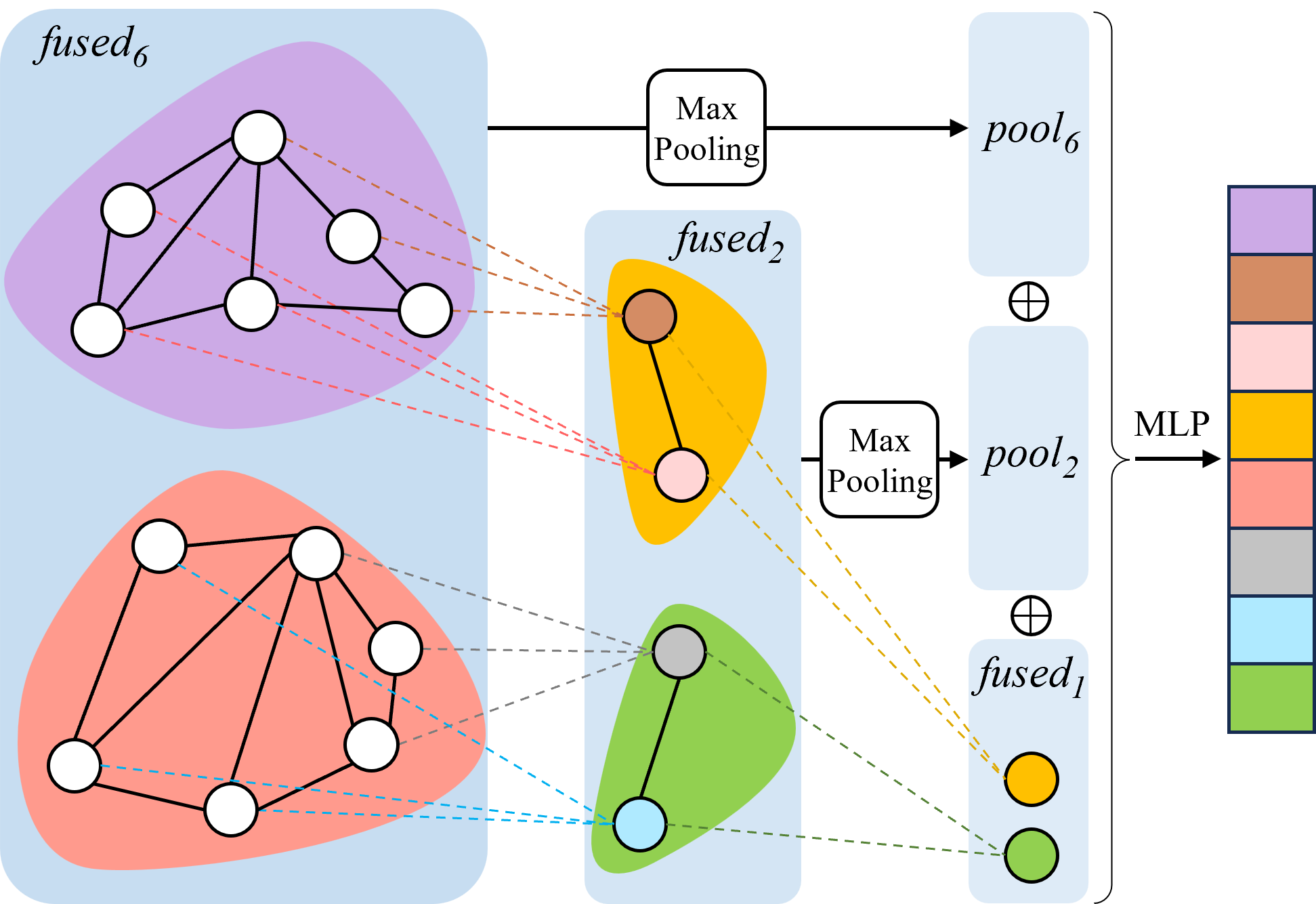}
\caption{Structure of the GHAM. Hierarchical pooling reduces the graph from 6 to 2, then to 1 node, with adaptive weighted fusion dynamically balancing modality features across scales.}
\label{diffpool}  
\end{figure}

To dynamically integrate LiDAR and RGB features, we propose the graph hierarchical abstraction module (GHAM), as depicted in Fig.~\ref{diffpool}, which retains the \textit{pooling GNN} while omitting the \textit{embedding GNN}. GHAM leverages the time-aware features from the TG-GAT, specifically $h_{l} = F_{lt}^6$ and $h_{r} = F_{rt}^6$ as input embeddings. The structure employs hierarchical pooling to enable progressive temporal abstraction, where six frame-level nodes are first grouped into two short-term clusters and then compressed into a single global node. This design allows GHAM to capture both short-term motion dynamics and long-term temporal dependencies, followed by adaptive weighted fusion across multiple scales to dynamically balance modality features. It consists of two main steps.

\subsubsection{Hierarchical Pooling}
For each modality, the first pooling layer collapses nodes $h_{l}, h_{r}$ into soft clusters $X_{l2}, X_{r2} \in \mathbb{R}^{2 \times 256}$, along with updated adjacency matrices $A_{l2}, A_{r2}$
    \begin{equation}
    S_1 = \text{softmax}(GNN_1(h, edge\_index))
    \end{equation}
    \begin{equation}
    \quad X_2 = S_1^T h
    \end{equation}
    \begin{equation}
    \quad A_2 = S_1^T A S_1
    \end{equation}
where $S_1 \in \mathbb{R}^{6 \times 2}$ is the assignment matrix, $GNN_1$ is the first \textit{pooling GNN}, $h \in \{h_{l}, h_{r}\}$, and $X_2 \in \{X_{l2}, X_{r2}\}$, $A_2 \in \{A_{l2}, A_{r2}\}$, $edge\_index$ is a $2 \times N$ tensor ($N$ is the number of edges), and $A$ is the adjacency matrix derived from $edge\_index$.

The second pooling layer further reduces $X_{l2}, X_{r2}$ to $X_{l1}, X_{r1} \in \mathbb{R}^{1 \times 256}$, along with updated adjacency matrices $A_{l1}, A_{r1}$
    \begin{equation}
    S_2 = \text{softmax}(GNN_2(X_2, A_2))
    \end{equation}
    \begin{equation}
    \quad X_1 = S_2^T X_2
    \end{equation}
    \begin{equation}
    \quad A_1 = S_2^T A_2 S_2
    \end{equation}
where $S_2 \in \mathbb{R}^{2 \times 1}$, $GNN_2$ is the second \textit{pooling GNN}, and $X_1 \in \{X_{l1}, X_{r1}\}$, and $A_1 \in \{A_{l1}, A_{r1}\}$.

\subsubsection{Weighted Fusion}
To balance the features of LiDAR and RGB modalities across different scales, we introduce learnable weights $\alpha_1, \alpha_2, \alpha_3 \in [0,1]$, initialized at 0.5 and constrained by a sigmoid function. The features are fused at the 6-node, 2-node, and 1-node stages as follows

\begin{equation}
    fused_6 = \alpha_1 h_l + (1 - \alpha_1) h_r
\end{equation}
where $fused_6 \in \mathbb{R}^{6 \times 256}$. Max-pooling is applied to generate $pool_6 \in \mathbb{R}^{1 \times 256}$.

\begin{equation}
    fused_2 = \alpha_2 X_{l2} + (1 - \alpha_2) X_{r2}
\end{equation}
where $fused_2 \in \mathbb{R}^{2 \times 256}$. Max-pooling is applied to obtain $pool_2 \in \mathbb{R}^{1 \times 256}$.

\begin{equation}
    fused_1 = \alpha_3 X_{l1} + (1 - \alpha_3) X_{r1}
\end{equation}
where $fused_1 \in \mathbb{R}^{1 \times 256}$.

The multi-scale features are combined via element-wise addition
\begin{equation}
    F_{rl} = pool_6 + pool_2 + fused_1.
\end{equation}

The $F_{rl}$ represents a hierarchically aggregated context that combines temporal and multi-modal information. This feature representation is directly fed into the RL state input, where it serves as the context for the decision-making policy in UGV navigation.

\section{Experiments} 
In this section, we demonstrate the superiority of our approach for UGV navigation in both simulated and real crowded environments via extensive comparisons and ablation experiments.

\subsection{Setups}
\subsubsection{Simulated Environments}
The DRL-TH framework with a multi-modal fusion scheme is trained in CARLA version 0.9.12, which provides highly realistic and crowded scenarios. We utilized the ROS bridge to enable the distributed collection of the UGV’s proprioceptive states and the visualization of local environmental observations. The training process consists of two curriculum stages: conventional scenarios and crowded obstacle scenarios. We used a vehicle measuring 2.2 m in length and 1.5 m in width as the training agent. We selected Town02 as the training scenario, while Town01, Town03, and Town07 were used as test scenarios. All experiments were conducted on a computer equipped with an Intel i9-13900KF CPU and an NVIDIA GeForce RTX 4090 GPU.

\begin{figure}[t]
\centering
\includegraphics[width=0.45\textwidth]{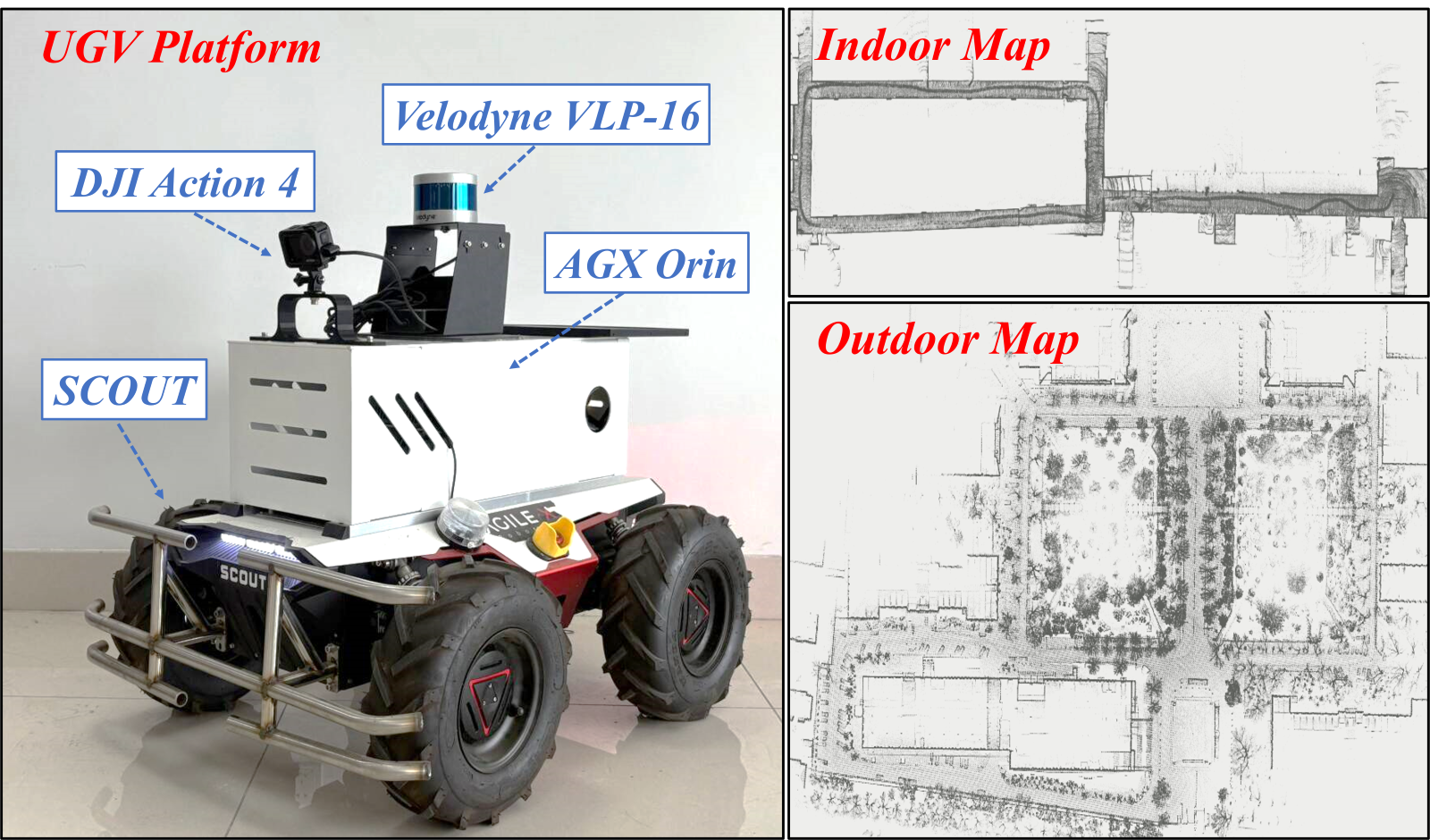}
\caption{The UGV and map used in the real-world test, where the main external sensors and configuration are annotated.}
\label{ugv}  
\end{figure}

\subsubsection{Real-World Environments}
When we conducted experiments in the real world, as shown in Fig. \ref{ugv}, a DJI Osmo Action 4 RGB camera with a height of 1.2 m, and a downward tilt of \SI{10}{\degree} is mounted on an AgileX SCOUT UGV. The UGV was also equipped with a WTGAHRS1 IMU, a VLP-16 LiDAR, and a NVIDIA AGX Orin. We collected 2,100 raw RGB images at 5 Hz and 4,500 raw point clouds at 5 Hz via remote control. To evaluate the generalization ability of DRL-TH, the test environment significantly differed from the scenarios used to collect RGB images for training ERFNet and point clouds for training CADNet. Note that the angular velocity output of the model was reduced by half to ensure safe navigation during indoor experiments.

\begin{table*}[h]
    \centering
    \caption{Global Settings for Two Tasks in Four Towns}
    \renewcommand{\arraystretch}{1}
    \label{task12}
    \begin{tabular}{c|cccc|cccc}
        \toprule
        \multirow{2}{*}{\textbf{Scenes}} & \multicolumn{4}{c|}{\textbf{Task 1}} & \multicolumn{4}{c}{\textbf{Task 2}} \\
        & {\textbf{Pedestrians}} & {\textbf{Cyclists}} & {\textbf{Vehicles}} & {\textbf{Average length (m)}} & {\textbf{Pedestrians}} & {\textbf{Cyclists}} & {\textbf{Vehicles}} & {\textbf{Average length (m)}} \\
        \cmidrule(lr){1-9}
        \textbf{Town02} & 5 & 2 & 5 & 127.32 & 46 & 8 & 12 & 184.01\\
        \textbf{Town01} & 6 & 2 & 5 & 150.85 & 60 & 12 & 18 & 241.12\\
        \textbf{Town03} & 12 & 4 & 9 & 207.69 & 75 & 17 & 23 & 301.58\\
        \textbf{Town07} & 5 & 2 & 5 & 110.64 & 38 & 8 & 10 & 150.79\\
        \bottomrule
    \end{tabular}
\end{table*}

\subsubsection{Implementation Details}
The configurations below are empirically optimized to achieve a balance between convergence speed, generalization ability, and computational efficiency. To ensure reproducibility, we detail the training settings for each module as follows:
\begin{itemize}
    \item \textbf{RGB}: We crop the images to a resolution of $84 \times 84$ and process them using ERFNet to generate binary segmentation images in the FPV. The model is trained for 600 epochs with batch size 64. We employ the ADAM optimizer with an initial learning rate of 0.001, which is reduced to 0.0001. The weight decay is set to 0.0002.
    \item \textbf{Point Cloud}: We retain points within a \SI{15}{\meter} range from the UGV’s center during training. These are processed by CADNet to generate binary segmentation images with a resolution of $64 \times 64$ in the BEV. We employ the ADAM optimizer with a learning rate of 0.001. The weight decay is set to 0.0005.
    \item \textbf{PPO}: We use the ADAM optimizer with a learning rate of 0.0001. The discount factor for reward computation is set to 0.99, and the generalized advantage estimation (GAE) discount factor is 0.95. The training batch size is 256, the entropy coefficient is 0.01, and the clipping value for the PPO algorithm is 0.1.
\end{itemize}

\subsection{Comparison Methods}
In our experiments, to verify the effectiveness of the proposed method, we also introduced two variants, DRL-T and DRL-H, derived from the DRL-TH. To ensure a fair comparison in the following ablation studies, the experimental settings for the variants DRL-T and DRL-H were kept identical to those of DRL-TH in terms of reward design and PPO hyperparameters during both training and testing.
\begin{itemize}
    \item \textbf{DRL-TH}: This is the complete version of our proposed method, with the network architecture illustrated in Fig. \ref{pipline}.
    \item \textbf{DRL-T}: This ablation variant retains all modules except GHAM. The TG-GAT output is reduced the dimensionality via an MLP and concatenated with the current LiDAR and RGB features to complete subsequent steps.
    \item \textbf{DRL-H}: This ablation variant retains all modules except TG-GAT, initializing the GHAM input with a GNN.
\end{itemize}

Additionally, we compared our method with with three baselines: MMRL~\cite{huang2021ral}, 
RGL~\cite{chen2020relational}, and ALVO~\cite{xieyd2025iros}.

\subsection{Evaluation in Simulated Environments}
We evaluated the above baselines using five widely adopted metrics from the navigation literature \cite{xie2023drl-vo, zhao2025rakd, huang2021ral}:
\begin{itemize}
    \item \textbf{Success Rate (SR)}: The fraction of collision-free trials.
    \item \textbf{Average Time (AT)}: The average travel time.
    \item \textbf{Average Linear Velocity (ALV)}: The average travel linear velocity.
    \item \textbf{Average Angular Velocity (AAV)}: The average travel angular velocity.
    \item \textbf{Path Smoothness (PS)}: The geometric smoothness of the path.
\end{itemize}

\begin{table*}[h]
    \centering
    \caption{Quantitative Comparisons of Different Methods in Simulated Environments Across Two Tasks}
    \renewcommand{\arraystretch}{1}
    \setlength{\tabcolsep}{10pt}
    \label{result-sim}
    \begin{tabular}{c|c|ccccc|ccccc}
        \toprule
        \multirow{2}{*}{\textbf{Scenes}} & \multirow{2}{*}{\textbf{Methods}} & \multicolumn{5}{c|}{\textbf{Task 1}} & \multicolumn{5}{c}{\textbf{Task 2}} \\
        & & {\textbf{SR}} & {\textbf{AT}} & {\textbf{ALV}} & {\textbf{AAV}} & {\textbf{PS}} & {\textbf{SR}} & {\textbf{AT}} & {\textbf{ALV}} & {\textbf{AAV}} & {\textbf{PS}} \\
        \cmidrule(lr){1-12}
        \multirow{4}{*}{\textbf{Town02}}
        &\textbf{RGL\cite{chen2020relational}} & 11/20 & 95.15 & 1.47 & 0.36 & 0.098 & 7/20 & 152.64 & 1.32 & 0.39 & 0.241\\
        &\textbf{MMRL\cite{huang2021ral}} & 14/20 & 88.42 & 1.63 & \textbf{0.11} & 0.023 & 9/20 & 130.96 & 1.55 & 0.24 & 0.093\\       
        &\textbf{ALVO\cite{xieyd2025iros}} & 18/20 & \textbf{78.34} & \textbf{1.83} & 0.13 & 0.054 & 16/20 & 116.18 & 1.73 & \textbf{0.17} & 0.132\\
        &\textbf{DRL-TH (Ours)} & \textbf{19/20} & 79.62 & 1.81 & 0.14 & \textbf{0.021} & \textbf{18/20} & \textbf{113.94} & \textbf{1.76} & 0.19 & \textbf{0.069}\\
        \cmidrule(lr){1-12}
        \multirow{4}{*}{\textbf{Town01}}
        &\textbf{RGL\cite{chen2020relational}} & 10/20 & 113.34 & 1.44 & 0.32 & 0.116 & 7/20 & 189.40 & 1.36 & 0.45 & 0.231\\
        &\textbf{MMRL\cite{huang2021ral}}& 14/20 & 104.63 & 1.56 & 0.18 & 0.027 & 8/20 & 168.39 & 1.53 & 0.32 & 0.108\\
        &\textbf{ALVO\cite{xieyd2025iros}} & 16/20 & 90.26 & 1.77 & \textbf{0.12} & 0.041 & 15/20 & \textbf{144.58} & \textbf{1.78} & 0.29 & 0.184\\
        &\textbf{DRL-TH (Ours)} & \textbf{18/20} & \textbf{88.72} & \textbf{1.85} & 0.13 & \textbf{0.023} & \textbf{17/20} & 149.54 & 1.72 & \textbf{0.28} & \textbf{0.074}\\
        \cmidrule(lr){1-12}
        \multirow{4}{*}{\textbf{Town03}}
        &\textbf{RGL\cite{chen2020relational}} & 6/20 & 147.93 & 1.51 & 0.41 & 0.107 & 1/20 & 304.50 & 1.10 & 0.49 & 0.367\\
        &\textbf{MMRL\cite{huang2021ral}} & 12/20 & 134.23 & 1.64 & \textbf{0.14} & 0.032 & 6/20 & 219.05 & 1.51 & 0.33 & 0.198\\        
        &\textbf{ALVO\cite{xieyd2025iros}} & 13/20 & 114.15 & \textbf{1.86} & 0.15 & 0.044 & 11/20 & 196.25 & 1.68 & 0.29 & 0.235\\
        &\textbf{DRL-TH (Ours)} & \textbf{17/20} & \textbf{117.20} & 1.87 & 0.17 & \textbf{0.031} & \textbf{15/20} & \textbf{190.42} & \textbf{1.69} & \textbf{0.27} & \textbf{0.079}\\
        \cmidrule(lr){1-12}
        \multirow{4}{*}{\textbf{Town07}}
        &\textbf{RGL\cite{chen2020relational}} & 10/20 & 88.65 & 1.41 & 0.39 & 0.101 & 6/20 & 124.77 & 1.37 & 0.44 & 0.238\\
        &\textbf{MMRL\cite{huang2021ral}} & 15/20 & 78.99 & 1.59 & 0.18 & 0.027 & 8/20 & 105.24 & 1.59 & 0.36 & 0.102\\         
        &\textbf{ALVO\cite{xieyd2025iros}} & 17/20 & \textbf{69.72} & 1.80 & \textbf{0.12} & 0.043 & 15/20 & 94.63 & 1.72 & \textbf{0.13} & 0.176\\
        &\textbf{DRL-TH (Ours)} & \textbf{19/20} & 70.57 & \textbf{1.83} & 0.14 & \textbf{0.024} & \textbf{17/20} & \textbf{91.79} & \textbf{1.80} & 0.14 & \textbf{0.072}\\
        \bottomrule
    \end{tabular}
\end{table*}

\begin{table}[t]
    \centering
    \caption{Comparison of Different Methods in Terms of Average Navigation Success Rate with Varying Pedestrian Densities}
    \renewcommand{\arraystretch}{1}
    \setlength{\tabcolsep}{6.0pt}
    \label{success_rate_table}
    \begin{tabular}{c|cccccc}
        \toprule
        \multirow{2}{*}{\textbf{Method}} & \multicolumn{6}{c}{\textbf{Normalized Pedestrian Density}} \\
        \cmidrule(lr){2-7}
        & {\textbf{0}} & {\textbf{0.1}} & {\textbf{0.2}} & {\textbf{0.3}} & {\textbf{0.4}} & {\textbf{0.5}} \\
        \cmidrule(lr){1-7}
        \textbf{RGL\cite{chen2020relational}} & 15/20 & 14/20 & 13/20 & 11/20 & 8/20 & 4/20\\
        \textbf{MMRL\cite{huang2021ral}} & 17/20 & 15/20 & 14/20 & 12/20 & 11/20 & 8/20\\
        \textbf{ALVO\cite{xieyd2025iros}} & 19/20 & 18/20 & 16/20 & 16/20 & 15/20 & 14/20\\
        \textbf{DRL-TH (Ours)} & 19/20 & 19/20 & 18/20 & 18/20 & 17/20 & 16/20\\
        \bottomrule
    \end{tabular}
\end{table}

\begin{figure}[t]
  \centering
  \subfloat[]{\includegraphics[width=0.48\columnwidth]{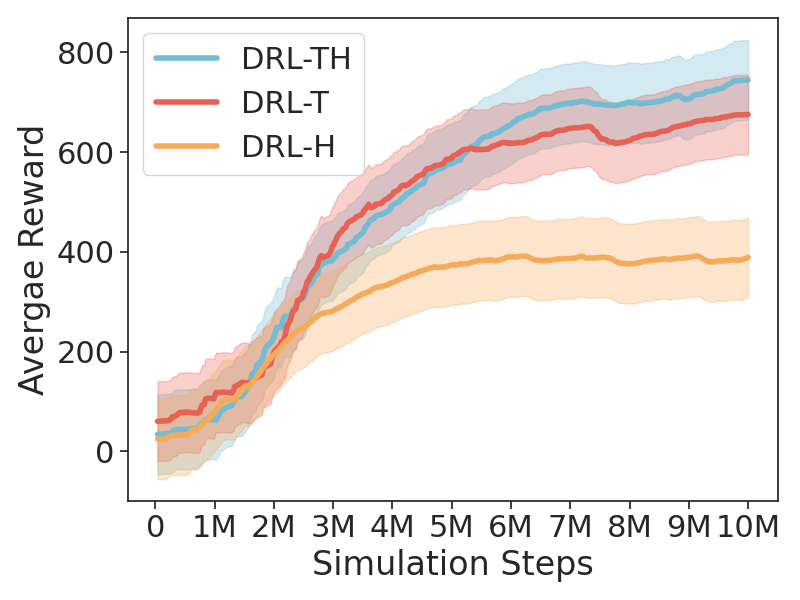}\label{subf1}}
  \hspace{0.01\columnwidth}
  \subfloat[]{\includegraphics[width=0.48\columnwidth]{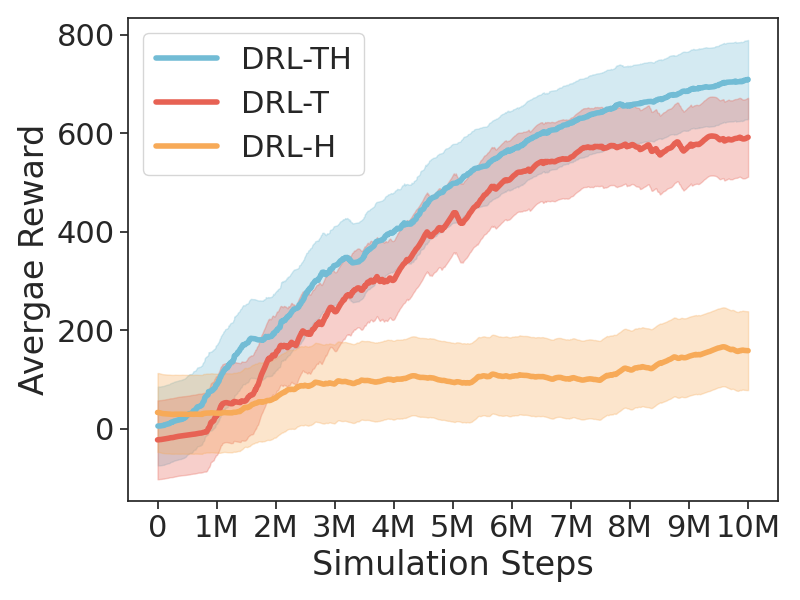}\label{subf2}}
  \caption{Average reward in two-stage training scenarios. (a) conventional stage. (b) crowded stage.}
  \label{combinedstage}
\end{figure}

\subsubsection{Reward Curves}
Fig. \ref{combinedstage} \subref{subf1} and \ref{combinedstage} \subref{subf2} illustrate the reward curves of DRL-TH, DRL-T, and DRL-H across two stages of the curriculum learning paradigm employed in our approach. The results demonstrate that DRL-TH has better rewards compared to the other baselines. This is due to the TG-GAT and GHAM, enabling DRL-TH to capture of scene evolution while dynamically adjusting the fusion ratio of LiDAR and RGB modalities. In contrast, without the influence of TG-GAT, DRL-H lacks the ability to perceive the future states of obstacles, leading to frequent collisions and impeding reward growth. Although DRL-T improves rewards via temporal context awareness, its inability to adaptively balance multi-modal features limits overall performance across different environments.

\subsubsection{Different Tasks}
We designed two tasks to evaluate the performance of all methods. Task 1 represents a simple conventional scenario, while Task 2 simulates a highly dynamic and crowded environment. The specific configurations are detailed in Tab. \ref{task12}. Each method was tested 20 times under both tasks, navigating from the starting point to the destination.

The results are presented in Tab. \ref{result-sim}. In Task 1, our DRL-TH achieved the highest success rates across the simple scenarios of four typical towns in CARLA. Notably, although DRL-TH did not excel in average velocity or average angular velocity, it demonstrated the best path smoothness among all methods. This performance was also observed in Task 2, which can be attributed to the path smoothness reward function introduced in Section~\ref{reward}. The MMRL performed well in Task 1 but exhibited a decline in Task 2. This is because the MMRL overly relies on image segmentation results, which fail when the visual field is saturated with a large number of obstacles, impacting the UGV navigation capability. The RGL underperformed in both tasks, due to its lack of long-range global path optimization and the need for frequent graph structure updates with numerous obstacles. Furthermore, accumulated prediction errors may lead to unstable paths, indirectly causing collisions or navigation failures. In contrast, ALVO delivered a suboptimal performance among all methods. Note that the limitations of its single-modal perception lead to the UGV exhibiting unstable motion when surrounded by obstacles in highly dynamic crowded environments.

\subsubsection{Different Obstacle Densities}
To evaluate the navigation performance of our DRL-TH framework in various crowded environments, we conducted experiments comparing DRL-TH with three methods under varying obstacle density levels. For consistency, all obstacles in this experiment were standardized as pedestrians. The pedestrian density is defined as $\rho_s = {N_p} / {A}$, where $\rho_s$ represents the pedestrian density (in pedestrians per square meter), $N_p$ is the number of pedestrians in the designated area, and $A$ is the area of the road segment. The experimental scenario was set in the Town02 map of the CARLA simulator, with a road segment of 30 m in length and 8 m in width, yielding a total area of $A = 240 \, \text{m}^2$. Based on real-world considerations, the road segment can accommodate a maximum of 30 pedestrians, resulting in a maximum density of 0.125. To facilitate comparison, the density was normalized as $\rho_n = \rho_s / 0.125$. We then conducted the experiments at various normalized density levels, with each tested 20 times and averaged to evaluate the navigation success rates and completion times of all methods. 

As pedestrian density increases, the navigation success rates of all methods exhibit a declining trend, shown in Tab. \ref{success_rate_table}. Notably, DRL-TH maintains a success rate of approximately 80\%, while the worst performing RGL achieves only a 40\% success rate when the normalized pedestrian density reaches 0.5. Furthermore, when pedestrian density is high, the success rates of MMRL and RGL decline more significantly. The primary reason is that both methods utilize 2D radar as their sensing modality, with its perception performance degrading under high pedestrian density, thus affecting the navigation success rate.

Additionally, the average navigation completion time of all methods increases to varying degrees as pedestrian density increases in Tab. \ref{average_time_table}. Even when the normalized pedestrian density reaches 0.5, the average time of DRL-TH is 19.1 seconds, which can be attributed to the velocity reward function design presented in Section~\ref{reward}. Compared to the other three methods, DRL-TH exhibits shorter completion times, due to they lack temporal reasoning, which limits them to reacting only to the current obstacle distribution, thereby delaying their navigation completion time.

\begin{table}[t]
    \centering
    \caption{Comparison of Different Methods in Terms of Average Navigation Completion Time with Varying Pedestrian Densities}
    \renewcommand{\arraystretch}{1}
    \setlength{\tabcolsep}{7.6pt}
    \label{average_time_table}
    \begin{tabular}{c|cccccc}
        \toprule
        \multirow{2}{*}{\textbf{Method}} & \multicolumn{6}{c}{\textbf{Normalized Pedestrian Density}} \\
        \cmidrule(lr){2-7}
        & {\textbf{0}} & {\textbf{0.1}} & {\textbf{0.2}} & {\textbf{0.3}} & {\textbf{0.4}} & {\textbf{0.5}} \\
        \cmidrule(lr){1-7}
        \textbf{RGL\cite{chen2020relational}} &21.3 &21.9 &22.6 &23.3 &23.8 &25.6\\
        \textbf{MMRL\cite{huang2021ral}} &19.5 &19.8 &20.7 &21.0 &21.6 &23.7\\   
        \textbf{ALVO\cite{xieyd2025iros}} &16.9 &17.2 &18.0 &18.8 &19.5 &20.4\\
        \textbf{DRL-TH (Ours)} &16.7 &17.1 &17.8 &18.2 &18.9 &19.1\\
        \bottomrule
    \end{tabular}
\end{table}

\subsubsection{Weighted Fusion}
\begin{figure}[t]
\centering
\includegraphics[width=0.45\textwidth]{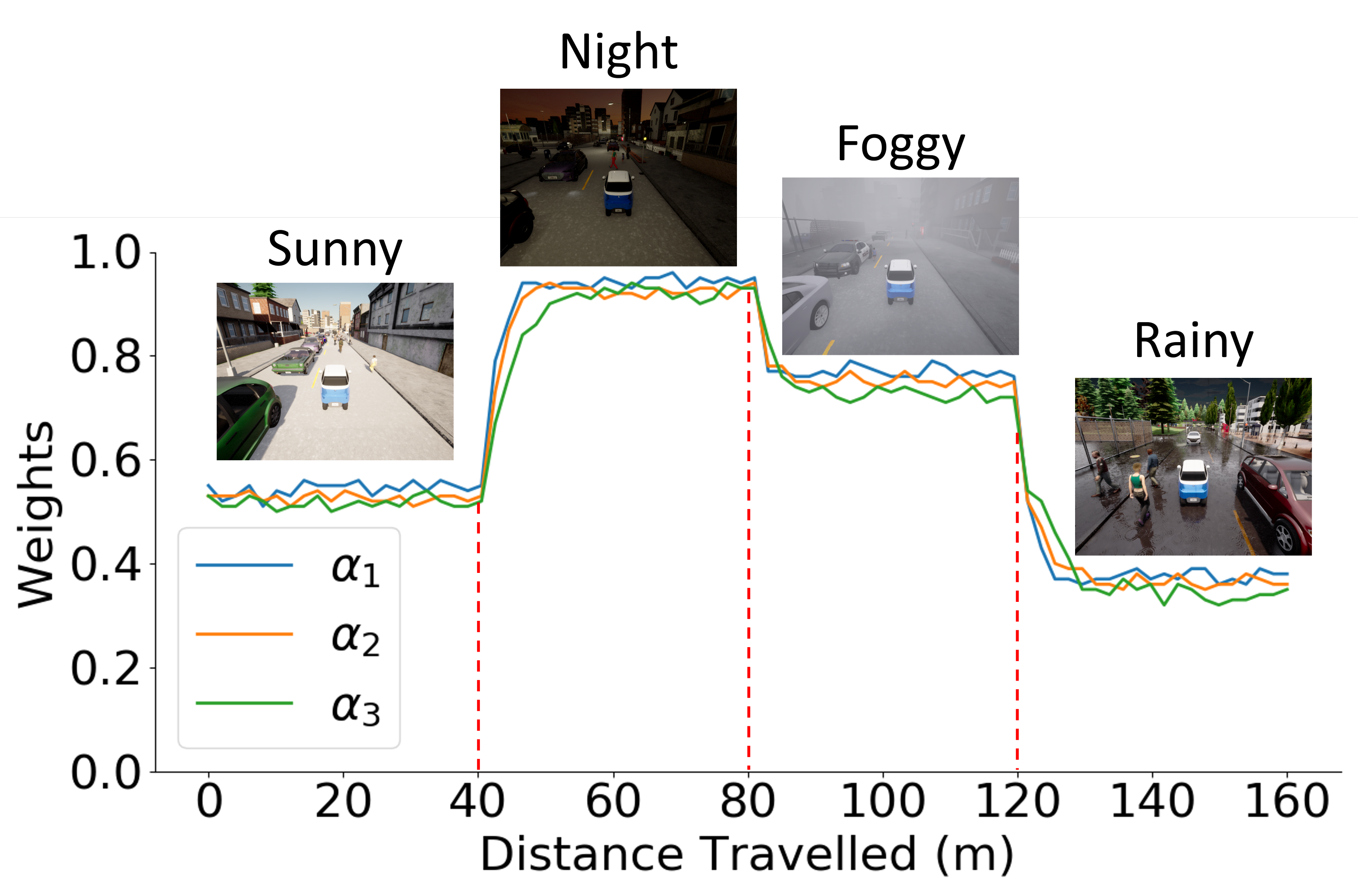}
\caption{Variation of the learnable weights $\alpha_1$, $\alpha_2$, and $\alpha_3$ of GHAM during navigation under different weather conditions.}
\label{a}  
\end{figure}

We also evaluate the dynamic variations of the learnable weights $\alpha_1$, $\alpha_2$, and $\alpha_3$ in GHAM as the UGV traverses four representative weather conditions. As illustrated in Fig. \ref{a}, in sunny, both LiDAR and RGB signals remain stable. When the scene shifts to night, RGB perception degrades significantly due to low illumination. In foggy, the RGB is significantly weakened by reduced contrast, while LiDAR still provides reliable geometric information at close distances. In rainy, LiDAR is affected by scattering, whereas RGB retains usable holistic appearance information. Overall, the temporal evolution of these weights demonstrates that GHAM can adaptively adjust the weights of each modality based on their instantaneous reliability under different environmental conditions.

\subsection{Evaluation in Real-World Environments}
Besides the simulation experiments, we also conducted a series of real-world experiments to validate the applicability of our DRL-TH policy. We focused on evaluating the UGV navigation performance of our system under low-density and high-density crowd scenarios. We tested in two different parts of campus, an indoor lobby, and an outdoor road environment, as shown in Fig. \ref{ugv}. In each environment, the UGV navigated through a series of predetermined subgoal points, with pedestrians or cyclists moving naturally in varying group sizes, prompting the UGV to dynamically adjust its path. The real UGV directly employed the DRL-TH policy trained on CARLA simulation platform without any fine-tuning.

\begin{figure*}[t]
    \centering
    \subfloat[]{%
        \includegraphics[width=0.24\textwidth]{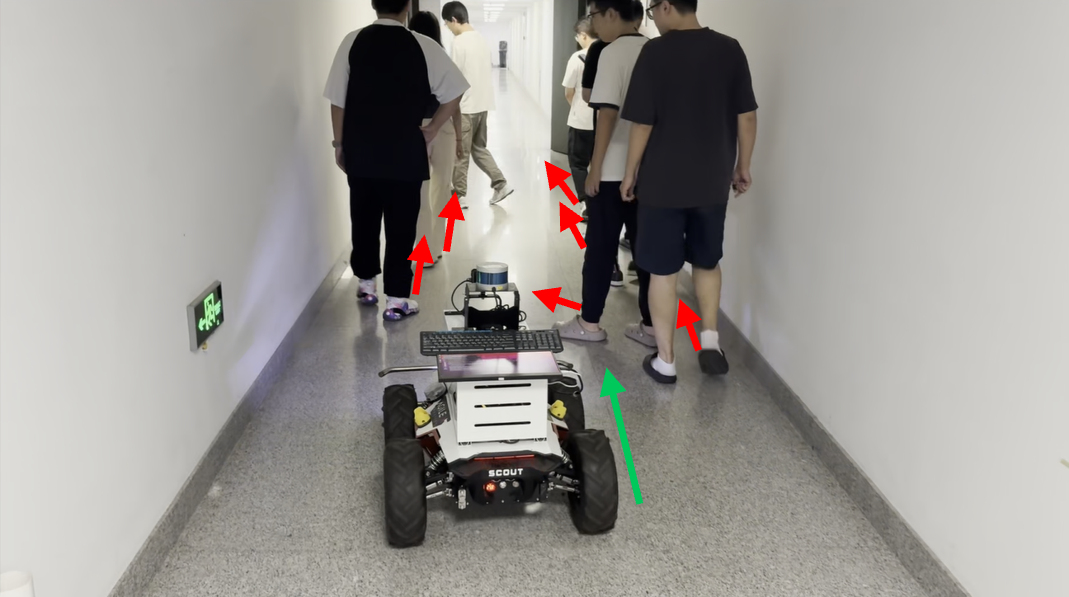}%
        \label{sub1}%
    }\hfill
    \subfloat[]{%
        \includegraphics[width=0.24\textwidth]{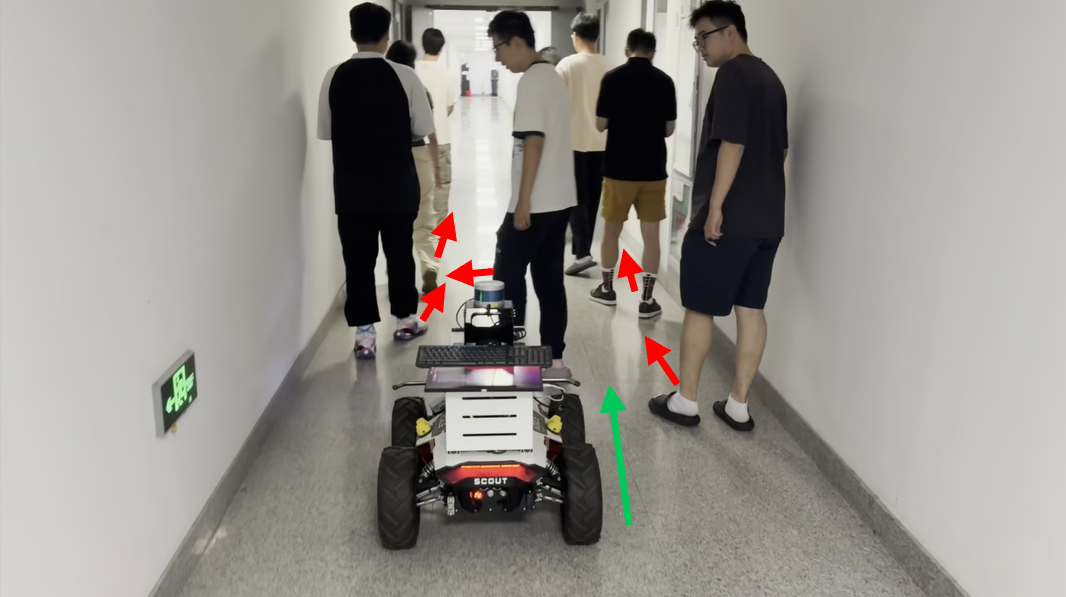}%
        \label{sub2}%
    }\hfill
    \subfloat[]{%
        \includegraphics[width=0.24\textwidth]{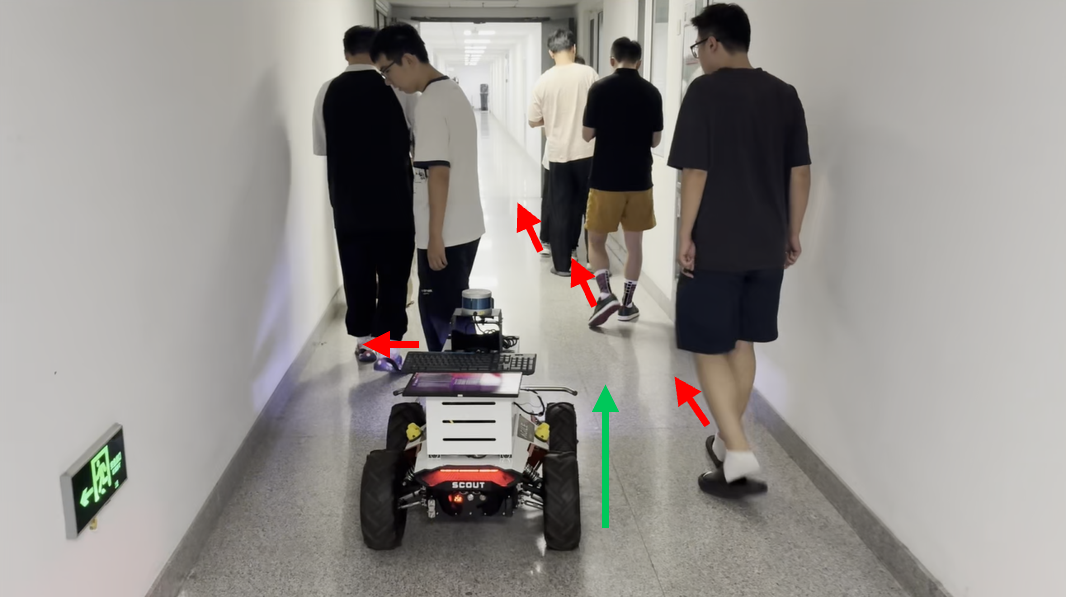}%
        \label{sub3}%
    }\hfill
    \subfloat[]{%
        \includegraphics[width=0.24\textwidth]{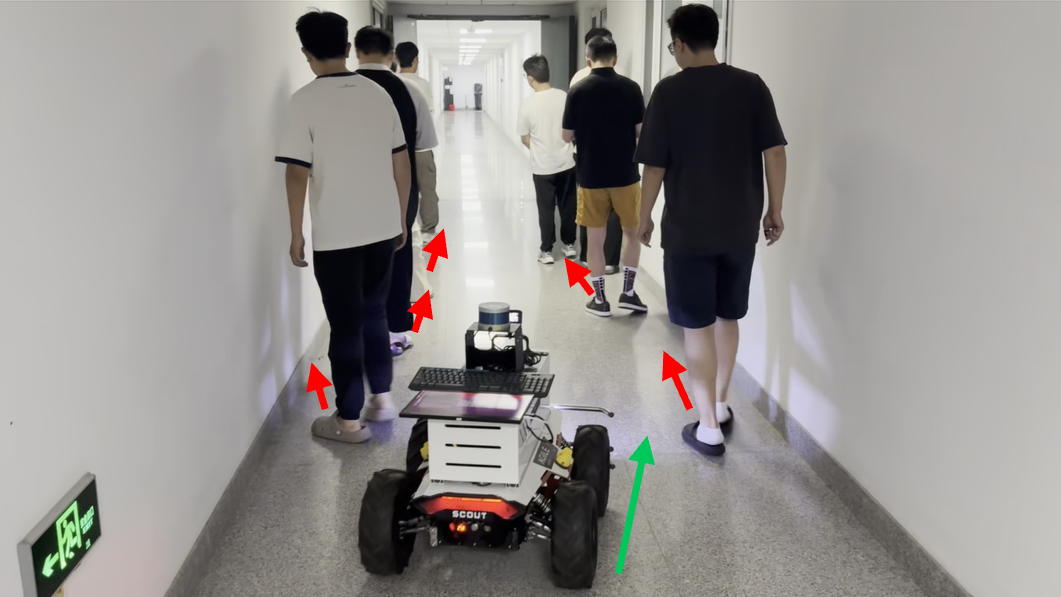}%
        \label{sub4}%
    }
    \caption{UGV reactions to moving pedestrians in the indoor environment with high crowd density at different times. (a) $t$. (b) $t+1$. (c) $t+2$. (d) $t+3$.}
    \label{i-high}
\end{figure*}

\begin{figure*}[t]
    \centering
    \subfloat[]{%
        \includegraphics[width=0.24\textwidth]{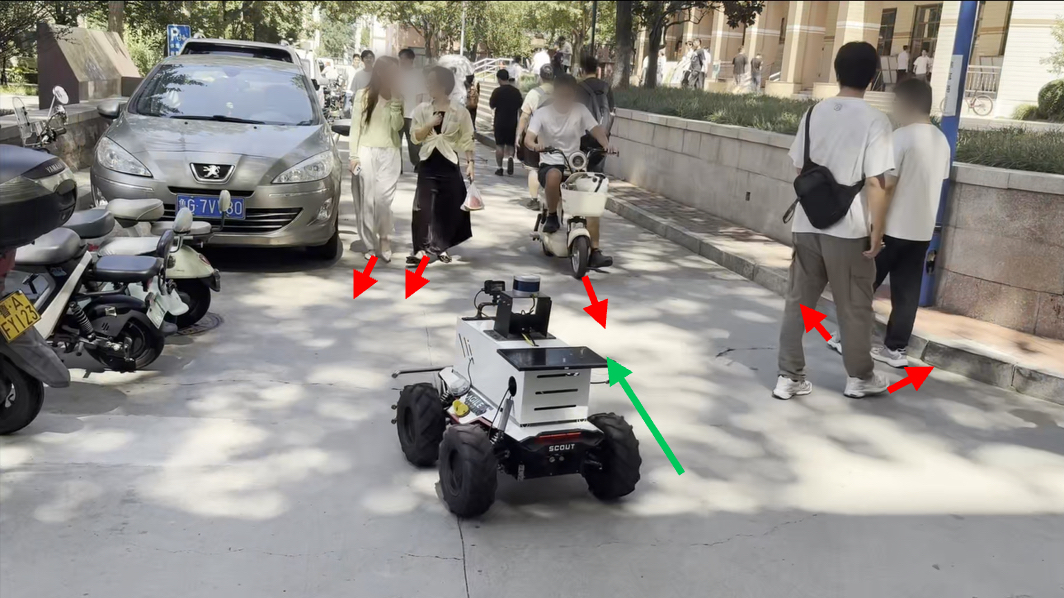}%
        \label{sub1}%
    }\hfill
    \subfloat[]{%
        \includegraphics[width=0.24\textwidth]{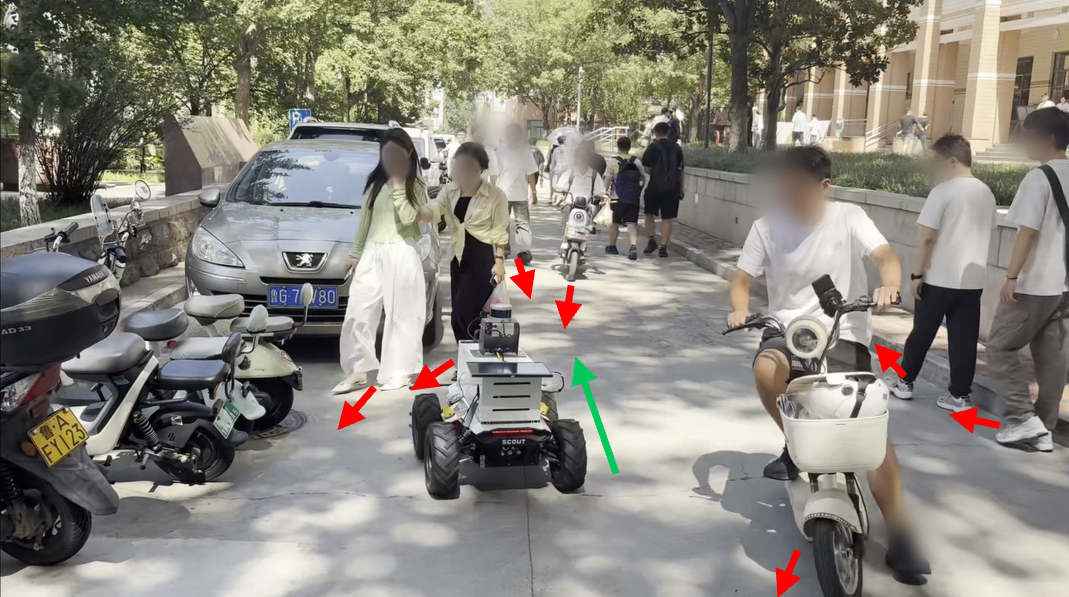}%
        \label{sub2}%
    }\hfill
    \subfloat[]{%
        \includegraphics[width=0.24\textwidth]{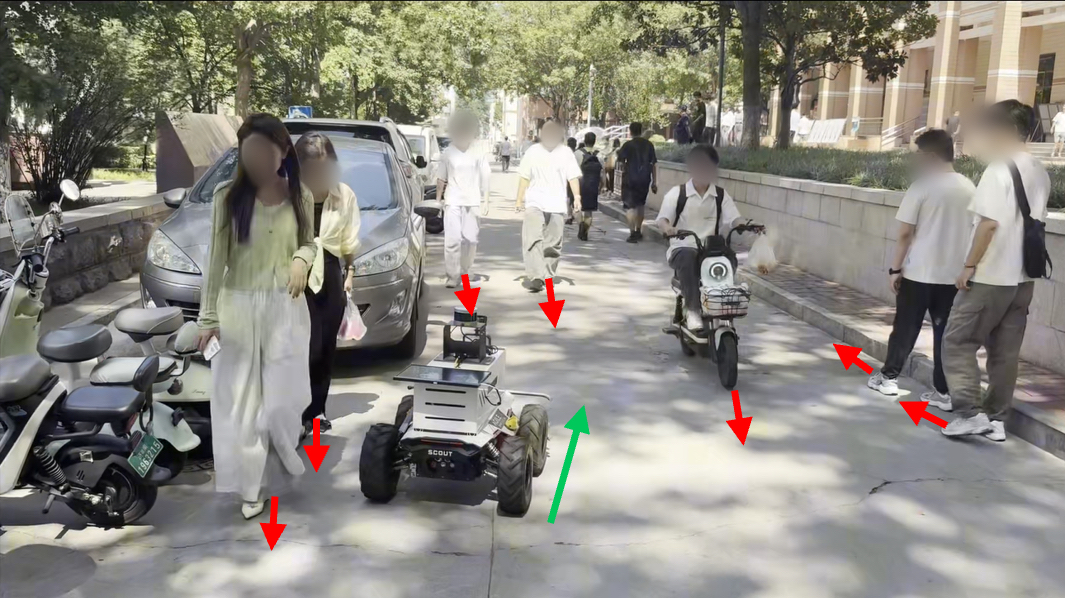}%
        \label{sub3}%
    }\hfill
    \subfloat[]{%
        \includegraphics[width=0.24\textwidth]{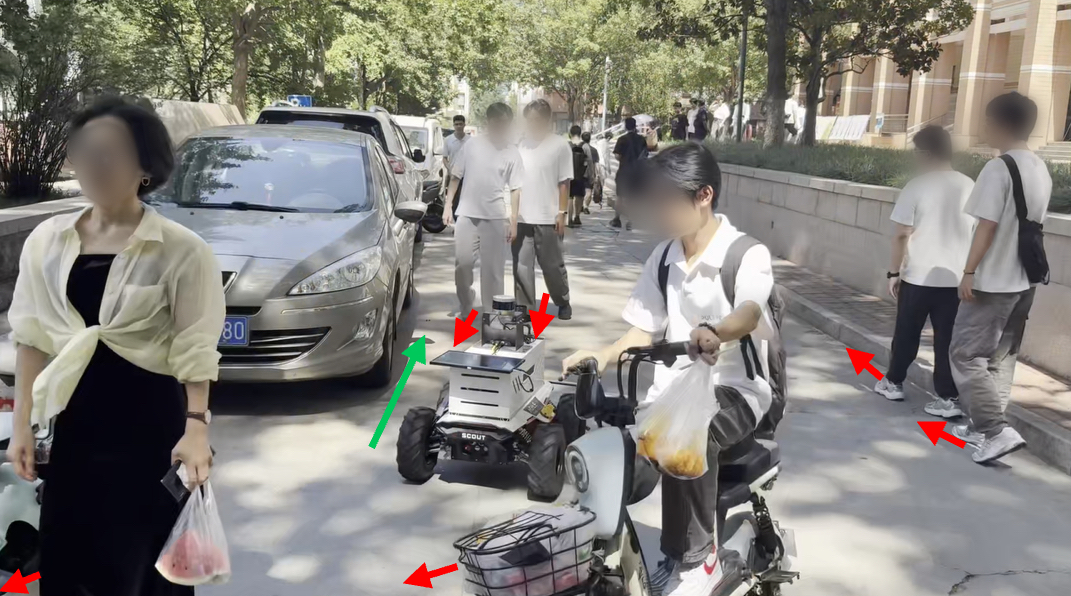}%
        \label{sub4}%
    }
    \caption{UGV reactions to moving pedestrians in the outdoor environment with high crowd density at different times. (a) $t$. (b) $t+1$. (c) $t+2$. (d) $t+3$.}
    \label{o-high}
\end{figure*}

We tested the UGV across various time periods to obtain performance results under different crowd densities. In low-density crowd scenarios, part 2 of the supplementary video illustrates how our UGV safely completed both indoor and outdoor navigation tasks in environments with sparse obstacles. It is apparent that in this case, the UGV could easily navigate its goals by tuning its forward direction toward subgoal points. In high-density crowd scenarios, Figs.~\ref{i-high} and ~\ref{o-high}, as well as the part 2 in the supplementary video present a more complex case compared to the former. Even in such a challenging dynamic environment, our UGV was still able to avoid collisions with pedestrians and cyclists, safely navigated around stationary pedestrians, and reached predefined goals. These results demonstrate that our DRL-TH policy is capable of generalizing to new environments with varying crowd densities.
We believe it primarily to a key factor, which involves preprocessing the data before inputting sensor data into our navigation strategy, transforming raw sensor inputs into intermediate-level feature representations. This reduces the gap between simulation and reality, allowing our policy learned entirely in simulation to work in the real world without any retraining or fine-tuning.

\section{Conclusion}
This paper proposes a DRL framework named DRL-TH, which leverages graph neural networks to enable smooth and safe navigation for UGVs in crowded environments. It comprises the temporal-guided graph attention network (TG-GAT) and graph hierarchical abstraction module (GHAM), which serve as two key components. TG-GAT enhances the ability of the UGV to capture of scene evolution by integrating temporal weights into attention scores. GHAM employs hierarchical pooling and learnable weighted fusion to dynamically integrate multi-modal features, feeding the results into a PPO network to output specific navigation commands. Extensive experiments were conducted in simulated and real-world scenes to demonstrate the effectiveness of the DRL-TH. For future work, we plan to integrate large models into the framework to replace handcrafted reward functions, enabling more adaptive reward shaping and enhancing the generalization capability of the navigation policy across unseen environments.

\bibliographystyle{IEEEtran}
\bibliography{export}

\end{document}